\documentclass[10pt,letterpaper]{article}

\usepackage{ccn}

\PassOptionsToPackage{hyphens}{url}\usepackage[hidelinks]{hyperref}

\usepackage{pslatex}
\usepackage{apacite}
\usepackage{graphicx,import}
\graphicspath{{./images/}}
\usepackage{subcaption}
\usepackage[shortlabels]{enumitem}
\usepackage{pgfplots}

\pgfplotsset{compat = 1.3}

\usepackage{soul}

\usepackage[eulergreek]{sansmath}

\usepackage[outercaption]{sidecap} 

\usepackage{tikz}
\usetikzlibrary{shapes,arrows}
\usetikzlibrary{positioning}
\usetikzlibrary{calc}
\usetikzlibrary{patterns}
\usepackage{tikzviolinplots}
\usepackage{tikz-network}
\usetikzlibrary{external}
\usetikzlibrary{plotmarks}

\newcommand{\xMarkerAsText}{%
	\begin{tikzpicture}[inner sep=0pt,baseline=-.5ex]%
		\draw[white](0,0) -- (1ex,1ex);%
		\node[mark size=3pt,color=black] at (0.5ex, 0.1ex){%
			\pgfuseplotmark{x}%
		};%
	\end{tikzpicture}%
}

\pgfplotsset{
	every axis label = {font=\sansmath\sffamily},
	legend style = {font=\sansmath\sffamily},
	label style = {font=\sansmath\sffamily}
}

\definecolor{mycolor1}{HTML}{D55E00}
\definecolor{mycolor2}{HTML}{56B4E9}
\definecolor{mycolor3}{HTML}{E69F00}
\definecolor{mycolor4}{HTML}{0072B2}
\definecolor{mycolor5}{HTML}{CC79A7}

\pgfplotscreateplotcyclelist{my-default-colors}{
	{mycolor1,fill=mycolor1!55!white},
	{mycolor2,fill=mycolor2!55!white},
	{mycolor3,fill=mycolor3!55!white},
	{mycolor4,fill=mycolor4!55!white},
	{mycolor5,fill=mycolor5!55!white},
}

\newlist{continuouslist}{enumerate*}{1}
\setlist[continuouslist]{label=\textbf{\roman*)},itemjoin=\ }

\title{Revealing an Unattractivity Bias in Mental Reconstruction of Occluded Faces using Generative Image Models}

\author{{\large \bf Frederik Riedmann (frederik.riedmann@fau.de)} \\
Departement Informatik
  \AND {\large \bf Bernhard Egger\footref{footnoteSharedAuthorship} (bernhard.egger@fau.de)} \\
  Departement Informatik
  \AND {\large \bf Tim Rohe\footref{footnoteSharedAuthorship} (tim.rohe@fau.de)} \\ %
  Institut für Psychologie   \vspace{1em} \\
  Friedrich-Alexander-Universität Erlangen-Nürnberg\\
Erlangen, Germany}

\begin{document}%
\maketitle%

\renewcommand*{\thefootnote}{\fnsymbol{footnote}}
\footnotetext[1]{These two authors contributed equally to this work.\label{footnoteSharedAuthorship}}
\renewcommand*{\thefootnote}{\arabic{footnote}}

\section{Abstract}
{
	\bf
	Previous studies have shown that faces are rated as more attractive when they are partially occluded. 
	The cause of this observation remains unclear. One explanation is a mental reconstruction of the occluded face parts which is biased towards a more attractive percept as shown in face-attractiveness rating tasks.
	We aimed to test for this hypothesis by using a delayed matching-to-sample task, which directly requires mental reconstruction. In two online experiments, we presented observers with unattractive, neutral or attractive synthetic reconstructions of the occluded face parts using a state-of-the-art diffusion-based image generator. Our experiments do not support the initial hypothesis and reveal an unattractiveness bias for occluded faces instead. This suggests that facial attractiveness rating tasks do not prompt reconstructions. Rather, the attractivity bias may arise from global image features, and faces may actually be reconstructed with unattractive properties when mental reconstruction is applied.
}
\begin{quote}
	\small
	\textbf{Keywords:} 
	Human Face Perception; Face Occlusion; Mental Reconstruction; Generative Models; Stable Diffusion
\end{quote}

\section{Intro and Research Question}
During the COVID-19 pandemic, medical masks not only helped to prevent infections, but also increased the wearer's perceived attractiveness: Numerous studies have shown that observers rate human faces as more attractive when they are partially occluded \cite{Orghian2020Jan, Patel2020Aug, Kamatani2021May, Pazhoohi2022Dec, Hies2022Dec}.  A recent hypothesis is that observers reconstruct occluded face parts using an averaged face template based on the faces they have seen previously in their life. Especially, for unfamiliar faces from other races than their own, this template would be skewed towards attractive faces (e.g. by celebrities), resulting in an attractivity bias \cite{Kamatani2023Jan}.

Previous studies used a simple task to measure the attractiveness bias: Participants were presented with a face with partial occlusion (e.g., superimposed mask or cropped image) or a full face and rated its overall attractiveness. However, this face-attractiveness rating (FAR) task does not require observers to mentally reconstruct the occluded face part, but observers provide holistic attractiveness ratings which may be biased by a general positive judgment of face masks (e.g., indicating responsible health behavior). In our study, we designed a delayed matching-to-sample (DMTS, \cite{Paule1998Sep}) task that requires observers to mentally reconstruct the occluded face part based on the visible face features: After a brief delay, observers matched one masked or unmasked face (sample face) to unmasked comparison faces with either attractive, neutral or unattractive mouth area of the face (match faces). The delay required observers to recollect the sample face from memory including a mental reconstruction of the occluded face part to perform the matching operation. Even though we could replicate the attractivity bias in the standard FAR task, our results from the DMTS task show that observers rather reconstruct the missing face part with an unattractive version, revealing an unattractivity bias in mental reconstruction of the missing face parts.

\section{Methods}
\begin{figure*}
	\begin{minipage}[c]{0.71\textwidth}
		\def\svgwidth{\textwidth}
		\fontsize{9pt}{1em} \begingroup%
		\makeatletter%
		\providecommand\color[2][]{%
			\errmessage{(Inkscape) Color is used for the text in Inkscape, but the package 'color.sty' is not loaded}%
			\renewcommand\color[2][]{}%
		}%
		\providecommand\transparent[1]{%
			\errmessage{(Inkscape) Transparency is used (non-zero) for the text in Inkscape, but the package 'transparent.sty' is not loaded}%
			\renewcommand\transparent[1]{}%
		}%
		\providecommand\rotatebox[2]{#2}%
		\newcommand*\fsize{\dimexpr\f@size pt\relax}%
		\newcommand*\lineheight[1]{\fontsize{\fsize}{#1\fsize}\selectfont}%
		\ifx\svgwidth\undefined%
		\setlength{\unitlength}{481.79497753bp}%
		\ifx\svgscale\undefined%
		\relax%
		\else%
		\setlength{\unitlength}{\unitlength * \real{\svgscale}}%
		\fi%
		\else%
		\setlength{\unitlength}{\svgwidth}%
		\fi%
		\global\let\svgwidth\undefined%
		\global\let\svgscale\undefined%
		\makeatother%
		\begin{picture}(1,0.6038798)%
			\lineheight{1}%
			\setlength\tabcolsep{0pt}%
			\put(0,0){\includegraphics[width=\unitlength,page=1]{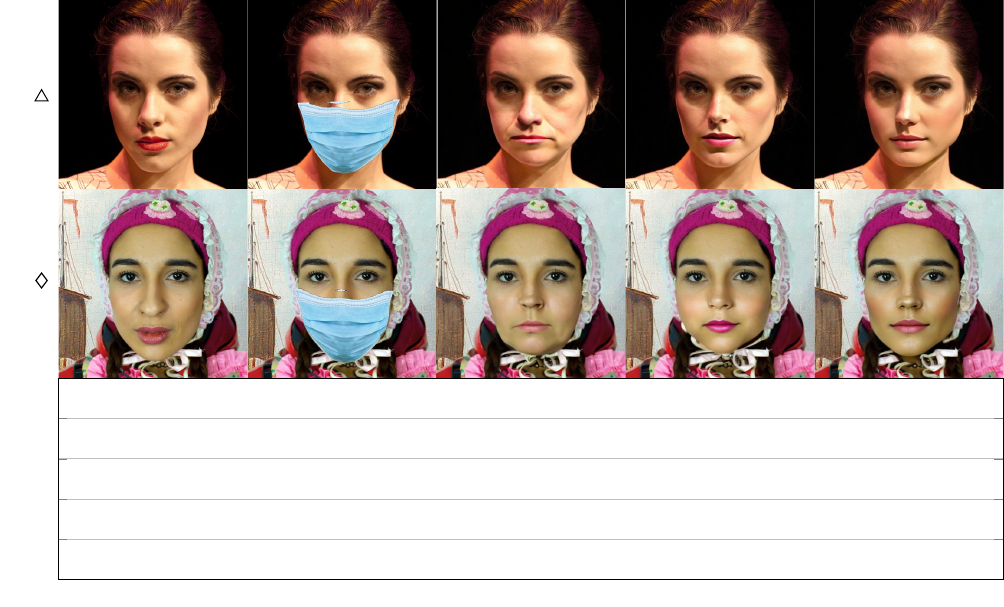}}%
			\put(0.03479221,0.01979833){\color[rgb]{0,0,0}\makebox(0,0)[lt]{\lineheight{1.25}\smash{\begin{tabular}[t]{l}0\end{tabular}}}}%
			\put(0.0244547,0.06002292){\color[rgb]{0,0,0}\makebox(0,0)[lt]{\lineheight{1.25}\smash{\begin{tabular}[t]{l}20\end{tabular}}}}%
			\put(0.0244547,0.10024751){\color[rgb]{0,0,0}\makebox(0,0)[lt]{\lineheight{1.25}\smash{\begin{tabular}[t]{l}40\end{tabular}}}}%
			\put(0.0244547,0.1404721){\color[rgb]{0,0,0}\makebox(0,0)[lt]{\lineheight{1.25}\smash{\begin{tabular}[t]{l}60\end{tabular}}}}%
			\put(0.0244547,0.18069662){\color[rgb]{0,0,0}\makebox(0,0)[lt]{\lineheight{1.25}\smash{\begin{tabular}[t]{l}80\end{tabular}}}}%
			\put(0.01411717,0.22092121){\color[rgb]{0,0,0}\makebox(0,0)[lt]{\lineheight{1.25}\smash{\begin{tabular}[t]{l}100\end{tabular}}}}%
			\put(0.00892869,0.02518029){\color[rgb]{0,0,0}\rotatebox{90}{\makebox(0,0)[lt]{\lineheight{1.25}\smash{\begin{tabular}[t]{l}Attractiveness ratings\end{tabular}}}}}%
			\put(0.10290447,0.00275582){\color[rgb]{0,0,0}\makebox(0,0)[lt]{\lineheight{1.25}\smash{\begin{tabular}[t]{l}(a) Original\end{tabular}}}}%
			\put(0.29102889,0.00287623){\color[rgb]{0,0,0}\makebox(0,0)[lt]{\lineheight{1.25}\smash{\begin{tabular}[t]{l}(b) Masked\end{tabular}}}}%
			\put(0,0){\includegraphics[width=\unitlength,page=2]{figure1.pdf}}%
			\put(0.46296164,0.00287623){\color[rgb]{0,0,0}\makebox(0,0)[lt]{\lineheight{1.25}\smash{\begin{tabular}[t]{l}(c) Unattractive\end{tabular}}}}%
			\put(0,0){\includegraphics[width=\unitlength,page=3]{figure1.pdf}}%
			\put(0.67016433,0.00287623){\color[rgb]{0,0,0}\makebox(0,0)[lt]{\lineheight{1.25}\smash{\begin{tabular}[t]{l}(d) Neutral\end{tabular}}}}%
			\put(0,0){\includegraphics[width=\unitlength,page=4]{figure1.pdf}}%
			\put(0.84993367,0.00287623){\color[rgb]{0,0,0}\makebox(0,0)[lt]{\lineheight{1.25}\smash{\begin{tabular}[t]{l}(e) Attractive\end{tabular}}}}%
			\put(0,0){\includegraphics[width=\unitlength,page=5]{figure1.pdf}}%
		\end{picture}%
		\endgroup%

	\end{minipage}%
	\hfill%
	\begin{minipage}[c]{0.27\textwidth}
		\caption{
			All five variations of exemplary identities. Violin plots display the distribution of all identities' mean attractiveness ratings. The two examples are highlighted in the violin plots and \protect\xMarkerAsText{} marks the mean value. %
			Original and masked images provided by the MaskedFace-Net \protect\cite{maskedfacenet2020} (CC BY-NC-SA 4.0).
		} \label{img:finalImageExamples}
	\end{minipage}
\end{figure*}

For our experiments, we created five different images for 100 face identities:
\begin{continuouslist}
	\item Original image
	\item Masked version, i.e. original image with a medical face mask covering mouth and nose
	\item Unattractive version with replaced oral area of low attractiveness
	\item Neutral version with replaced oral area of neutral attractiveness
	\item Attractive version with replaced oral area of high attractiveness.
\end{continuouslist}
Examples can be seen in Figure \ref{img:finalImageExamples}.

The original and masked versions were taken from the MaskedFace-Net dataset \cite{maskedfacenet2020}. To create the unattractive, neutral and attractive version, the same area that was occluded in the masked version was inpainted using the ``stable-diffusion-2-inpainting'' model \cite{Rombach_2022_CVPR}. Multiple prompts for faces with a neutral expression and low or high attractiveness were embedded, and a linear extrapolation between these vectors was used to generate images of different levels of attractiveness.
All prompts were structurally similar to following examples, but words were replaces with synonyms or increases:
\begin{continuouslist}
	\item ``an attractive beautiful person with neutral facial expression''
	\item ``a good looking person with a perfect face and neutral facial expression''
	\item ``an ugly person with neutral facial expression''
	\item ``a hideous appalling person with neutral facial expression''.
\end{continuouslist}
In addition to common negative prompts like ``injuries, unhealthy, mutated body, bad anatomy, deformed, bad composition, disfigured'', we also discouraged ``open mouth, emotion, smile'' and ``frown''.

Using these images, we performed two online surveys on SoSci Survey to conduct the following experiments \cite{soscisurvey}:
\begin{continuouslist}
	\item In Experiment~1, we implemented a standard FAR task as a replication and control: Participants were presented with all five versions of each image and rated the faces' overall attractiveness ratings using a slider (mapped to a 100-point scale). %
	\item In Experiment~2, we implemented a DMTS task using two experimental conditions: The participants were shown the masked version or the unmasked original image as a sample face for 1.25 seconds. This image disappeared and after a delay of 0.5 seconds, the three unattractive, neutral or attractive inpainted face images appeared in random spatial order.  Participants were asked to select one face as a match for which they believed it was most similar to the sample face. For the masked faces, this task required participants to mentally reconstruct the missing face part.  The unmasked sample face served as a control condition which also required recall of the sample face from memory, but no mental reconstruction of the missing part because the face was fully visible. Thus, comparing to this condition controlled for memory processes during recall of the sample face and response biases when choosing the sample face. This enabled us to isolate the effects of mental reconstruction. Further, each individual observer was randomly presented with each individual face (n = 99 trials in total) only in either masked or unmasked version to avoid retesting effects.
\end{continuouslist}

After giving written informed consent, 126 participants completed survey~1 (81 male, 43 female, 2 other; mean age 25.2 \textbf{±} 8.2 STD years) and 41 participants completed survey~2 (21 female, 20 male; mean age 34.4 \textbf{±} 17.3 STD years).  All images were re-categorized into the unattractive, neutral and attractive categories based on participants' attractiveness ratings in all subsequent analyses. 

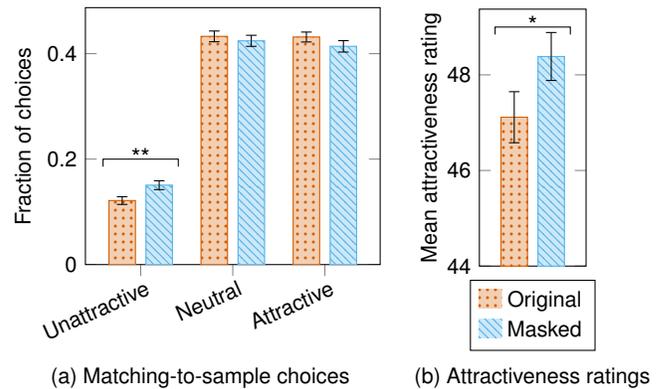
\begin{figure}
	\centering
	\begin{subfigure}[b]{.6\columnwidth}
		
		\pgfplotsset{tick label style={font=\small\sansmath\sffamily}}
		
		\begin{tikzpicture}
			\begin{axis}[
				ybar,
				enlarge x limits = 0.3,
				bar width = 10pt,
				ybar   = 4pt,
				ylabel={Fraction of choices},
				ylabel style={font=\small},
				legend style={at={(.5,-.5)},anchor=south,legend columns=1,font=\small},
				symbolic x coords={Unattractive, Neutral, Attractive},
				xtick=data,
				xtick pos=left,
				x tick label style={rotate=24,anchor=east,xshift=4pt,yshift=-5pt,font=\small},
				ymin = 0,
				height=50mm,
				width=55mm,
				cycle list name=my-default-colors,
				]
				\addplot+[
				preaction={fill, mycolor1!30!white},
				pattern = dots,
				pattern color=mycolor1,
				error bars/.cd,
				y dir=both,
				y explicit,
				error bar style={color=black}
				] coordinates {
					(Unattractive, 0.1214) +- (0, 0.0073) 
					(Neutral, 0.4328) +- (0, 0.0102)
					(Attractive, 0.4316) +- (0, 0.0094)
				};
				
				\addplot+[
				preaction={fill, mycolor2!30!white},
				pattern = north west lines,
				pattern color=mycolor2,
				error bars/.cd,
				y dir=both,
				y explicit,
				error bar style={color=black}
				] coordinates {
					(Unattractive, 0.1506) +- (0, 0.0084) 
					(Neutral, 0.4244) +- (0, 0.0105)
					(Attractive, 0.4139) +- (0, 0.0108)
				};
				
				\pgfmathsetmacro{\yvalue}{0.2}
				\pgfmathsetmacro{\spanWidth}{5mm}
				\draw ([yshift=-1mm, xshift=-\spanWidth]axis cs:Unattractive,\yvalue) -- ([xshift=-\spanWidth]axis cs:Unattractive,\yvalue) -- node[above, yshift=-2mm]{**} ([xshift=\spanWidth]axis cs:Unattractive,\yvalue) -- ([yshift=-1mm, xshift=\spanWidth]axis cs:Unattractive,\yvalue);
				
			\end{axis}
		\end{tikzpicture}

		\caption[Matching-to-sample choices]{Matching-to-sample choices}
		\label{figure:variationChoices}
	\end{subfigure}%
	\hfill%
	\begin{subfigure}[b]{.37\columnwidth}
		\pgfplotsset{tick label style={font=\small\sansmath\sffamily}}
		
		\begin{tikzpicture}
			\begin{axis}[
				ybar,
				enlarge x limits = 0.3,
				bar width = 10pt,
				ybar   = 4pt,
				ylabel={Mean attractiveness rating},
				ylabel style={font=\small},
				legend style={at={(.5,-.33)},anchor=south,legend columns=1, font=\small},
				ylabel shift = -4pt,
				symbolic x coords={Mean},
				xtick=\empty,
				xtick pos=left,
				ymin = 44,
				height=50mm,
				width=30mm,
				cycle list name=my-default-colors,
				legend image code/.code={%
					\draw[#1, draw=none] (0cm,-0.1cm) rectangle (0.3cm,0.15cm);
				},
				]
				\addplot+[
				preaction={fill, mycolor1!30!white},
				pattern = dots,
				pattern color=mycolor1,
				error bars/.cd,
				y dir=both,
				y explicit,
				error bar style={color=black}
				] coordinates {
					(Mean, 47.114) +- (0, 0.5338)
				};
				
				\addplot+[
				preaction={fill, mycolor2!30!white},
				pattern = north west lines,
				pattern color=mycolor2,
				error bars/.cd,
				y dir=both,
				y explicit,
				error bar style={color=black}
				] coordinates {
					(Mean, 48.384) +- (0, 0.5017) 
				};
				
				\pgfmathsetmacro{\yvalue}{49}
				\pgfmathsetmacro{\spanWidth}{5mm}
				\draw ([yshift=-1mm, xshift=-\spanWidth]axis cs:Mean,\yvalue) -- ([xshift=-\spanWidth]axis cs:Mean,\yvalue) -- node[above, yshift=-2mm]{*} ([xshift=\spanWidth]axis cs:Mean,\yvalue) -- ([yshift=-1mm, xshift=\spanWidth]axis cs:Mean,\yvalue);
				
				\legend{Original, Masked}
			\end{axis}
		\end{tikzpicture}

		\caption[Attractiveness ratings]{Attractiveness ratings}
		\label{figure:positivityBias}
	\end{subfigure}%
	\caption[Survey Results]{Survey results: Error bars indicate standard error.}%
\end{figure}

\section{Results}
The FAR task in experiment~1 showed that we were able to replicate the previous attractiveness bias for masked over unmasked original faces (one-sided t-test, $t_{2027} = 1.893$, $p = 0.029$; see Figure~\ref{figure:positivityBias}). Critically, our DMTS task in experiment~2 revealed the opposite result, that is an unattractiveness bias for masked faces (Figure~\ref{figure:variationChoices}): For each participant and masked versus unmasked sample face, we computed the fraction of trials in which either the unattractive, neutral or attractive face was chosen as a match. A repeated-measures ANOVA with factors sample-face masking (yes vs. no) and match-face version (unattractive vs. attractive) showed that participants generally preferred the attractive over the unattractive match face (main effect, $F_{1, 125} = 659.73$, $p < 0.001$).   Yet, for masked as compared to unmasked sample faces, participants more often chose the unattractive but less often chose the attractive match face (interaction effect, $F_{1, 125} = 5.48$, $p=0.02$). Two-sided t-tests confirmed that participants chose the unattractive version more often for a masked sample ($t_{125} = 2.98$, $p = 0.004$). This was not the case for the attractive versions ($t_{125} = -1.26$, $p = 0.208$) or the neutral match faces ($t_{125} = -0.61$, $p = 0.545$).  

A possible explanation for this effect could be that the participants select the images only by low-level perceptual similarity (i.e., independent from mental reconstruction) and that the unattractive images are on average more similar to the sample face. Control analyses showed that this is not the case: The choice probabilities of the three variations do not correlate significantly with their average distance (as measured with ArcFace \cite{arcface}) to the original face. This shows that the participants similarity judgement differs from that of ArcFace. Furthermore, the attractive faces have an average distance to the original face of 0.4548, as compared to 0.5669 for the unattractive faces. So while the unattractive faces are less similar to the original face, participants still chose them more often when the sample face was masked. This control analysis supports our finding that an unattractiveness bias arises from participants’ mental reconstruction

\section{Conclusion}
Our experiments replicated that face masks create an attractivity bias in overall attractiveness ratings. In contrast to this established finding \cite{Orghian2020Jan, Patel2020Aug, Kamatani2021May, Pazhoohi2022Dec, Hies2022Dec}, our DMTS task revealed an unattractivity bias for masked faces. Because the DMTS task requires observers to mentally ``fill-in'' the masked part, this result contradicts the idea that observers mentally reconstruct occluded face parts by using an attractive average face template. 
Instead, observers mentally reconstructed the face with unattractive features. Control analyses showed that this effect could not be explained by perceptual similarities between match and sample face.
The previously reported attractiveness bias may result from general factors that make a masked face appear more attractive in overall attractiveness rating tasks. For example, masks may increase perceived attractiveness by indicating  responsible health behavior or high social status of medical staff. Further, an attractiveness rating task can be performed without mentally reconstructing occluded face parts as hypothetically assumed. In conclusion, by combining a DMTS task with a state-of-the-art inpainting model, we show that the classical attractiveness bias does not not arise from mental reconstruction because this cognitive operation leads to an unattractivity bias instead.

\newpage
\bibliographystyle{apacite}
\setlength{\bibleftmargin}{.125in}
\setlength{\bibindent}{-\bibleftmargin}
\bibliography{ccn_style}

\begin{thebibliography}{}
\bibitem [\protect \citeauthoryear {%
Pazhoohi%
\ \BBA {} Kingstone%
}{%
Pazhoohi%
\ \BBA {} Kingstone%
}{%
{\protect \APACyear {2022}}%
}]{%
Pazhoohi2022Dec}
\APACinsertmetastar {%
Pazhoohi2022Dec}%
\begin{APACrefauthors}%
Pazhoohi, F.%
\BCBT {}\ \BBA {} Kingstone, A.%
\end{APACrefauthors}%
\unskip\
\newblock
\APACrefYearMonthDay{2022}{{\APACmonth{12}}}{}.
\newblock
{\BBOQ}\APACrefatitle {{Unattractive faces are more attractive when the
  bottom-half is masked, an effect that reverses when the top-half is
  concealed}} {{Unattractive faces are more attractive when the bottom-half is
  masked, an effect that reverses when the top-half is concealed}}.{\BBCQ}
\newblock
\APACjournalVolNumPages{Cogn. Research}{7}{1}{1--10}.
\newblock
\begin{APACrefDOI} \doi{10.1186/s41235-022-00359-9} \end{APACrefDOI}
\PrintBackRefs{\CurrentBib}
\end{thebibliography}

\end{document}